\documentclass{opt2023} 

\usepackage{parskip}
\title[Information-Theoretic Trust Regions for Stochastic Gradient-Based Optimization]{Information-Theoretic Trust Regions for Stochastic Gradient-Based Optimization}

\optauthor{%
\Name{Philipp Dahlinger}$^1$ \Email{philipp.dahlinger@kit.edu}\\
\Name{Philipp Becker}$^{1, 2}$ \\
\Name{Maximilian Hüttenrauch}$^1$ \\
\Name{Gerhard Neumann}$^{1, 2}$ \\ 
\addr 1: Karlsruhe Institute of Technology, Germany \\
\addr 2: FZI Research Center for Information Technology, Germany}

\usepackage{graphicx}
\usepackage{bm}
\usepackage{optidef}
\usepackage{tikz}
\usetikzlibrary{calc}
\usepackage{multirow}
\usetikzlibrary{positioning}
\usepackage{wrapfig}

\usepackage[ruled]{algorithm2e}
\usepackage{physics}
\usepackage{booktabs}
\usepackage{subcaption}
\usepackage[font=small]{caption}
\usepackage[capitalise]{cleveref}

\newcommand{\ittr}{arTuRO}
\newcommand{\RR}{\mathbb{R}}

\DeclareMathOperator{\kl}{KL}
\DeclareMathOperator{\diag}{diag}

\usepackage{pgfplots}
\pgfplotsset{
    compat=1.17,
    /pgfplots/ybar legend/.style={
    /pgfplots/legend image code/.code={%
       \draw[##1,/tikz/.cd,yshift=-0.25em]
        (0cm,0cm) rectangle (3pt,0.8em);},},}

\usetikzlibrary{positioning} 
\usetikzlibrary{calc}
\usetikzlibrary{pgfplots.groupplots}
\usetikzlibrary{external}

\begin{document}
\maketitle
\begin{abstract}%
Stochastic gradient-based optimization is crucial to optimize neural networks. 
While popular approaches heuristically adapt the step size and direction by rescaling gradients, a more principled approach to improve optimizers requires second-order information. 
Such methods precondition the gradient using the objective's Hessian. 
Yet, computing the Hessian is usually expensive and effectively using second-order information in the stochastic gradient setting is non-trivial.
We propose using Information-Theoretic Trust Region Optimization (\ittr) for improved updates with uncertain second-order information.  
By modeling the network parameters as a  Gaussian distribution and using a Kullback-Leibler divergence-based trust region, our approach takes bounded steps accounting for the objective's curvature and uncertainty in the parameters. 
Before each update, it solves the trust region problem for an optimal step size, resulting in a more stable and faster optimization process.
We approximate the diagonal elements of the Hessian from stochastic gradients using a simple recursive least squares approach, constructing a model of the expected Hessian over time using only first-order information.
We show that \ittr{} combines the fast convergence of adaptive moment-based optimization with the generalization capabilities of SGD. 
\end{abstract}


\section{Introduction}

In this work, we introduce Inform\textbf{a}tion-Theo\textbf{r}etic \textbf{T}r\textbf{u}st \textbf{R}egion \textbf{O}ptimization (\ittr{}), an approach using approximate second-order information for stochastic optimization\footnote{Code available at \url{https://github.com/ALRhub/arturo}.}.
While using second-order information for preconditioning and step size selection is ubiquitous in classical optimization literature \citep{boyd2004convex}, it is not broadly adopted for stochastic optimization in deep learning as it suffers from two main problems. 
First, controlling the step size, using, e.g., a line search, breaks down if the second-order information is only approximate \citep{mokhtari2014res, bottou2018optimization}.
Second, while modern automatic differentiation tools allow computing the Hessian, this is often impractical due to high memory and processing demands. 

\ittr{} addresses the uncertainty in the stochastic second-order information by limiting the maximal update step using trust regions.
We take a distributional view of the network parameters and formalize a constrained optimization problem with an information-theoretic trust region \citep{peters2010relative, Abdolmaleki2015} that considers the objective's curvature.
Furthermore, we estimate the Hessian's diagonal using gradient information to avoid computing it explicitly. 
Using only a single gradient per step, we iteratively approximate the Hessian using recursive least squares and use a drift model \citep{Saerkkae_2013} to account for the parameters changing during updates. 
Assuming a Gaussian random walk of the Hessian, older gradients have a smaller weight in the regression.

We show that \ittr{} profits from both the trust regions and its improved way of estimating the Hessian on several standard benchmark tasks and model architectures where it exhibits similar or better performance compared to optimizers with moment-based adaptations and SGD.

\begin{wrapfigure}{r}{0.3\textwidth}
\vspace{-1cm}
        \resizebox{0.3\textwidth}{!}{\input{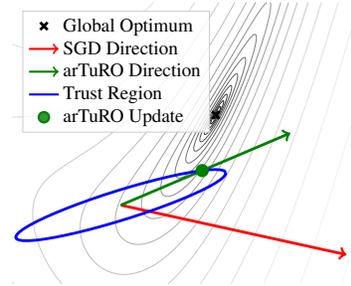}}
\caption{Visualization of the main idea behind \ittr{} - Information-Theoretic Trust Regions. 
        \ittr{}'s update direction (green arrow) is clearly better than simply stepping in the gradient direction (red).
        Yet, due to the stochastic nature of the optimization, fully relying on second-order information can be suboptimal. 
        Thus, we use a trust region (blue) based on the current parameter uncertainty to restrict the update length. }
\end{wrapfigure}


\section{Information-Theoretic Trust Region Optimization}
We study a typical supervised learning task where the aim is to solve an unconstrained non-convex optimization problem of the form
%
\begin{equation}
    \textstyle
    \min_{\bm \theta} \mathcal{L}(\bm \theta) = \frac{1}{N} \sum_{i=1}^N l_i(\bm \theta; x_i, y_i).
    \label{eq:full_batch_loss}
\end{equation}
%
Here, $\bm \theta \in \RR^n$ are the parameters of a neural network we aim to optimize, $(x_i, y_i)$ are pairs of input data and output targets while $l_i$ defines a differentiable loss function. The objective is usually approximated using mini-batches instead of the entire dataset.


\paragraph{Trust Region Optimization using Second-Order Information.}
We define a distribution over the parameters in order to introduce an information-theoretic trust region. A natural choice is a Gaussian distribution $   \pi_t(\bm \theta) = \mathcal{N}(\bm \theta; \bm \mu_t, \bm \Sigma_t)$, 
with mean $\bm \mu_t$ and covariance $\bm \Sigma_t$. The probabilistic view allows us to define a principled trust region.
Given a bound $\varepsilon \ge 0$, we bound the change in parameter distribution by the Kullback-Leibler (KL) divergence
$ \kl(\pi_t || \pi_{t-1}) \le \varepsilon$.
We can now solve \cref{eq:full_batch_loss} by repeatedly minimizing the expectation of the loss function over the distribution $\pi_t$ under the trust-region constraint
\begin{equation*}
    \min_{\pi_t} \mathbb{E}_{\bm\theta \sim \pi_t}[\mathcal{L}(\bm \theta)], \quad \text{s.t.}\,\kl(\pi_t || \pi_{t-1}) \le \varepsilon.
\end{equation*}
The primal solution of this problem has a closed-form solution for quadratic functions $\mathcal{L}(\bm \theta)$\citep{Abdolmaleki2015}. 
Thus, we use a second-order Taylor expansion of the loss function around the current mean $\bm \mu_t$ of the parameter distribution 
$   \mathcal{L}(\bm \theta) \approx f_t(\bm \theta) = 0.5\bm \theta^T \bm A_t \bm \theta + \bm \theta^T \bm b_t + c_t$.
Since the Hessian $\bm H_{\bm \mu_t}$ and the gradient $\nabla \mathcal{L}(\bm \mu_t)$ of the full dataset  is unknown due to mini-batching, we have a further approximation of the Taylor coefficients $\bm A_t \approx \bm H_{\bm \mu_t}$ and $\bm b_t \approx \nabla \mathcal{L}_t(\bm \mu_t) - \bm H_{\bm \mu_t} \bm \mu_t$.
This formulation allows for closed-form solutions of the expectation in the optimization problem as the distribution over parameters $\bm \theta$ is Gaussian.

\paragraph{Disentangled Trust Regions and Entropy Regularization.}
Previous work \citep{aboldmaleki2018MPO, huttenrauch2022regret} shows that disentangling the KL in independent constraints for the change of the mean and covariance improves the optimization procedure.
The KL divergence for Gaussian distributions has a closed-form solution $\kl(\pi_t || \pi_{t-1}) = C_\mu + C_\Sigma$
with
\begin{align*}
    C_\mu &= 0.5(\bm \mu_t - \bm \mu_{t-1})^T \bm \Sigma_{t-1}^{-1}(\bm \mu_t - \bm \mu_{t-1}) \quad \text{and}\\
    C_\Sigma &= 0.5\big[\tr(\bm \Sigma_{t-1}^{-1} \bm \Sigma_t) - n + \log \det(\bm \Sigma_{t-1}) - \log \det(\bm \Sigma_t) \big].
\end{align*}
As we are ultimately interested in a maximum a posteriori (MAP) solution, i.e., the mean of the parameter distribution, we propose to use the covariance part of the KL divergence in the \ittr{} objective instead of using it as a constraint.
This regularizes the optimal updated covariance towards the old covariance $\bm \Sigma_{t-1}$.
We further introduce a KL divergence $C_\lambda = \kl(\pi_t || \mathcal{N}(0 ,\lambda^{-1}\bm I))$ to a prior distribution and add it to the objective which regularizes the entropy of the distribution.
The resulting objective of the \ittr{} optimization problem is given as
\begin{align}
    \min_{\pi_t} \mathbb{E}_{\bm\theta \sim \pi_t}[f_t(\bm\theta)] + \rho C_\lambda + \nu C_\Sigma \quad \text{s.t.}~C_\mu \le \varepsilon. \label{obj:itropt}
\end{align}
The factors $\rho, \nu \in \RR$ weight the regularization, while the precision $\lambda \in \RR$ defines the prior's scaling.

\paragraph{Solving the Trust Region Problem.}\label{sec:main}
We use the method of Lagrange multipliers to find a solution to \cref{obj:itropt}.
Taking the derivative of the Lagrangian with respect to the primal parameters $\bm \mu_t$ and $\bm \Sigma_t$ and setting it to $0$, we obtain the primal solutions
\begin{align}
    \bm \mu_t(\eta) &= (\bm A_t + \eta \bm \Sigma_{t-1}^{-1} + \rho\lambda \bm I_n)^{-1}(\eta \bm \Sigma_{t-1}^{-1}\bm \mu_{t-1} - \bm b_t), \label{eq:primal_mean} \\
    \bm \Sigma_t &= (\rho + \nu)(\bm A_t + \rho \lambda \bm I_n + \nu \bm \Sigma_{t-1}^{-1})^{-1}\label{eq:primal_var}
\end{align}
where $\eta \ge 0$ is a Lagrange multiplier.
Thus, we see how $\eta$ interpolates between fully trusting and discarding the second-order information. 
To compute the optimal value $\eta^\ast$ we need to maximize the dual objective $g(\eta)$ corresponding to \cref{obj:itropt}.
For a detailed derivation, we refer to \cref{sec:solution_opt_problem}.
In order to scale to high dimensional problems we use a diagonal parameterization of the covariance $\bm \Sigma_t = \diag(\bm \sigma_t^2)$ and the matrix $\bm A_t = \diag(\bm a_t)$. 

It remains to define the update of the surrogate model. The general idea is to accumulate the previous gradient information of the loss function into the current update using recursive least squares with a drift model. The solution is a Kalman update which is scalable to high-dimensional problems. In \cref{sec:drift_model} we describe the derivation and the update equations of $\bm A_t$ and $\bm b_t$. 

A crucial part of \ittr{} is computing the Lagrange multiplier $\eta \ge 0$, which gives the step size of the update. 
Here, we use a bisection method due to its simplicity and amenability to a GPU implementation. 
As the change in $\eta$ between iterations is small, we can initialize the bisection using tight bounds around the previous $\eta$.
Using this method, we usually only need $2$ to $4$ iterations to find a sufficiently precise $\eta$.
For hyperparameters and further details, see \cref{sec:exp_details}.



\section{Experiments}

\begin{figure*}[t]
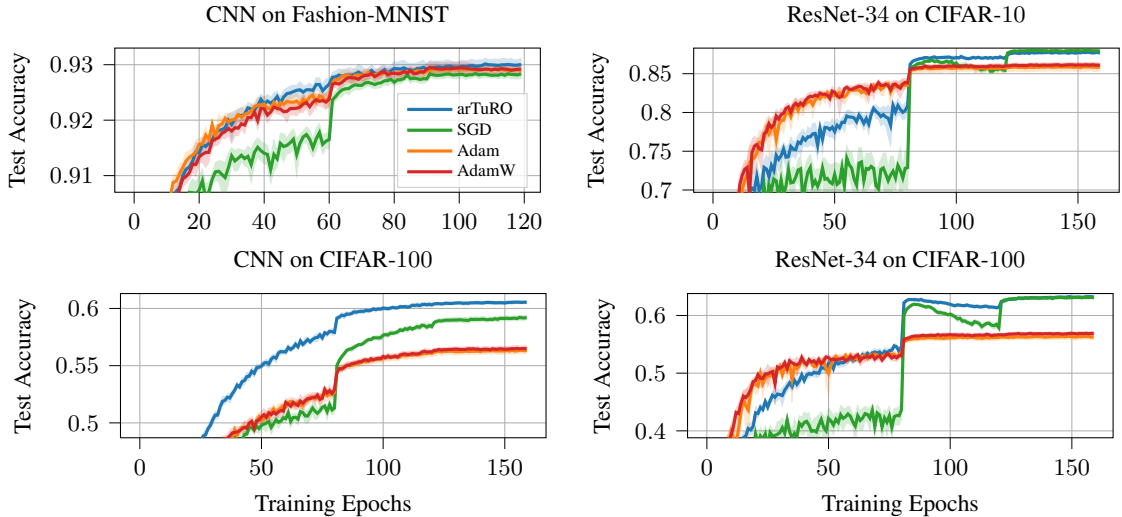

\begin{minipage}{0.5\textwidth}
   \centering
   \input{figures/results/fmnist_dropout_cnn_test_acc.tex}
\end{minipage}%
\begin{minipage}{0.5\textwidth}
    \centering
    \input{figures/results/cifar10_resnet34_test_acc}
\end{minipage}
\begin{minipage}{0.5\textwidth}
    \centering
    \input{figures/results/cifar100_dropout_cnn_test_acc}
\end{minipage}%
\begin{minipage}{0.5\textwidth}
    \centering
    \input{figures/results/cifar100_resnet34_test_acc}
\end{minipage}
\caption{Test accuracies of \ittr{}, SGD, Adam and AdamW.}
\label{fig:results}
\end{figure*}

We test the performance of \ittr{} on the well-established benchmarks Fashion-MNIST, CIFAR-$10$, and CIFAR-$100$. 
As the network architecture greatly affects the optimization procedure, we use a classical convolutional neural network (CNN), as well as ResNet architectures to evaluate the algorithms over different-sized networks. 
For architecture details, see \cref{app:exp_details}.
We repeat each experiment over $10$ seeds.
In the Fashion-MNIST CNN experiment, we train the network for $120$ epochs, while the other experiments run for $160$ epochs.

\begin{table}[t]
\centering
\resizebox{\textwidth}{!}{
    \begin{tabular}{lcccccc}\toprule
        & \multicolumn{2}{c}{\textbf{Fashion-MNIST}} & \multicolumn{2}{c}{\textbf{CIFAR-10}} &
        \multicolumn{2}{c}{\textbf{CIFAR-100}} \\
                  & CNN  & ResNet-$18$ & CNN & ResNet-$34$  & CNN & ResNet-$34$ \\ 
      \midrule 
      SGD & $92.84 \pm 0.08$ & $\bm{92.64} \pm \bm{0.07}$ & $87.11 \pm 0.15$ & $\bm{87.92} \pm \bm{0.16}$ & $59.20 \pm 0.18$ & $\bm{63.05} \pm \bm{0.16}$ \\ 
      Adam & $92.90 \pm 0.10$ & $91.76 \pm 0.17$ & $85.86 \pm 0.15$ & $85.89 \pm 0.51$ & $56.35 \pm 0.27$ & $56.32 \pm 0.45$ \\ 
      AdamW & $92.92 \pm 0.09$ & $91.97 \pm 0.11$ & $85.88 \pm 0.12$ & $86.07 \pm 0.30$ & $56.50 \pm 0.22$ & $56.86 \pm 0.28$ \\ 
      \midrule 
      \ittr{} & $\bm{93.01} \pm \bm{0.10}$ &  $92.13 \pm 0.12$ & $\bm{87.37} \pm \bm{0.13}$ & $87.66 \pm 0.19$ & $\bm{60.53} \pm \bm{0.13}$ & $\bm{63.19} \pm \bm{0.24}$ \\ 
      \bottomrule
    \end{tabular}}
\caption{Results of a CNN and a ResNet architecture on Fashion-MNIST, CIFAR-$10$, and CIFAR-$100$. We report the mean and the doubled standard error of the test accuracy in $[\%]$.} \label{tab:accuracy}
\end{table}
    
\paragraph{Results.} %
We compare \ittr{} against the state-of-the-art optimization algorithms SGD with momentum, Adam~\citep{Kingma2015}, and AdamW~\citep{LoshchilovH19}. All algorithms including \ittr{} use weight decay. We also include a learning rate scheduler for SGD, Adam, and AdamW, while in \ittr{}, we schedule the trust region bound $\varepsilon$. Twice during the optimization, the effective step size is reduced by a factor\footnote{Since the trust region bound of \ittr{} scales quadratically, we use a different scaling factor of $0.006$. Empirically, this results in the same reduction of the step size.} of $0.1$.
For a fair comparison, we conduct an extensive hyperparameter optimization for each objective and algorithm. In \cref{tab:accuracy}, we report the resulting final accuracy on previously unseen test data. 
\ittr{} performs superior on all CNN architectures and is comparable to SGD and clearly outperforms Adam and AdamW on the ResNet tasks.
Furthermore, SGD outperforms the adaptive learning rate methods Adam and AdamW in every task besides Fashion-MNIST CNN. \cref{app:exp_details} lists the used hyperparameters for each experiment and algorithm. In \cref{fig:results}, we present the accuracy on the test set throughout the optimization for four of the six experiments.
\ittr{} combines the faster convergence of Adam and AdamW with the generalization potential of SGD and benefits greatly from learning rate decay steps similar to SGD. Adam and AdamW can only compete on the CNN architecture on Fashion-MNIST (top-left). SGD on ResNet-$34$ shows a more unstable behavior as the test accuracy drops during a period of the optimization procedure. While SGD achieves a slightly higher final test accuracy, it is harder to tune.
The remaining two figures, as well as the train loss curves, are given in \cref{app:additional_results}.

\begin{figure}[t]
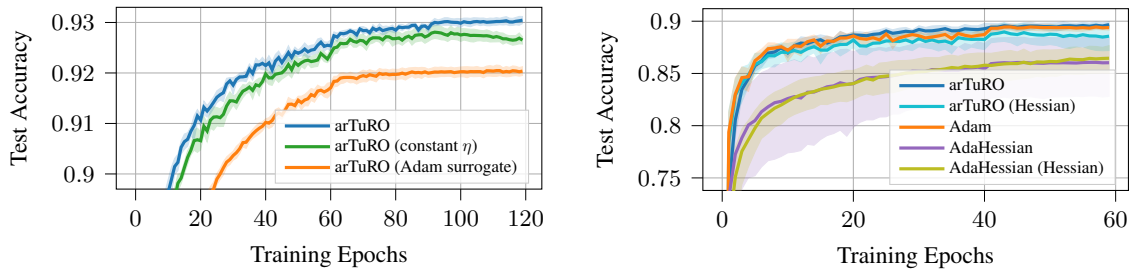

\begin{minipage}{0.5\textwidth}
\input{figures/ablation/trust_region_ablation_test_acc}
\end{minipage}
\begin{minipage}{0.5\textwidth}
\centering
\input{figures/ablation/hessian_ablation_test_acc.tex}
\end{minipage}%
\caption{Ablation Results. \textbf{Left:} \ittr{} ablation of the trust region and the surrogate computation. We report the test accuracy of the Fashion MNIST CNN task over the elapsed epochs. The blue line is the standard \ittr{} algorithm, the green line uses a constant $\eta$ instead of the solution to the dual problem, and the orange line replaces the recursive least squares surrogate computation with Adam's first and second moments estimation. \textbf{Right:} Ablation on Fashion MNIST with a smaller network where the computation of the analytic diagonal Hessian is feasible. \emph{\ittr{}(Hessian)} uses a running average over the Hessian and the gradient instead of our squared surrogate.
\emph{AdaHessian (Hessian)} uses the AdaHessian algorithm with the exact Hessian computation.}
\label{fig:ablation}
\end{figure}

\paragraph{Ablations.}%
First, we investigate the step size selection while using the identical update direction of the \ittr{} algorithm. We compare the trust region approach with an ablation where the dual parameter $\eta$ remains fixed. 
We carefully selected the fixed $\eta$ with insights from the trust region approach as a general choice of $\eta$ is not obvious. 
Additionally, we use a running average of the moment estimates from the Adam algorithm as the parameters of the quadratic surrogate to show that while the trust region approach is highly effective, it depends on an accurate model.
We present the results on the left side in \cref{fig:ablation}. 
We can conclude that a good surrogate is a key property for second-order algorithms. When having an adequate surrogate, the information-theoretic trust region gives an extra performance increase.

Second, we compare the second-order estimation of \ittr{} on a smaller network architecture against other second-order estimation techniques.
To this end, we compute the exact (stochastic) diagonal of the Hessian with a variation of the backpropagation algorithm \citep{pytorch_backpack}.
The running average of the Hessian and the gradient is then used as the surrogate parameters. 
We compare against Adam and the diagonal Hessian approximation of AdaHessian \citep{yao2020adahessian}.
To further evaluate AdaHessian approximation to the exact diagonal of the Hessian, we replace the AdaHessian approximation with the exact Hessian without altering the rest of the AdaHessian algorithm.
The results of the ablation displayed in \cref{fig:ablation} (right side) indicate that correct Hessian information based on mini-batches in the stochastic setting does not result in better optimization results and we can outperform other second-order methods with our fit of the quadratic surrogate.

\section{Related Work}

\paragraph{First-Order Methods.}
First-order optimization methods like Stochastic Gradient Descent (SGD) \citep{sgd1951} are widely used in deep learning for their simplicity and effectiveness. To improve performance, techniques like momentum~\citep{sutskever13, NESTEROV1983} and adaptive learning rates~\citep{Zeiler2012, Hinton2012} are have been introduced. The Adamalgorithm~\citep{Kingma2015} is a popular variant that employs both of them. Still, there are challenges, such as choosing the correct learning rate, addressing convergence issues~\citep{ReddiKK18}, and correctly including regularization~\citep{LoshchilovH19}. 
While variants of Adam, such as AMSGRAD~\cite{ReddiKK18} and AdamW~\cite{LoshchilovH19} try to address these issues, there are remaining issues with generalization~\cite{Keskar2017, WilsonRSSR17}, convergence~\cite{tan2019} and the theoretical understanding~\cite{zhou2020towards} of Adam and its variants.

\paragraph{Second-Order Methods.}
Using second-order information to precondition the gradients and automatically setting the learning rate has many practical and theoretical benefits in classical optimization \citep{AgarwalBH17,dry009}.
However, classical Hessian approximation schemes\cite{lbfgs} are not straightforwardly applicable in the stochastic optimization setting. 
While many approaches \cite{mokhtari2014res, yao2020adahessian} propose alternatives to approximate and use Hessians in this setting, devising a generally applicable, efficient second-order optimizer for large-scale stochastic optimization remains an open research question.

\paragraph{Information Theoretic Trust Regions.}
Information-theoretic trust regions find widespread use in reinforcement learning (RL)\citep{pmlr-v37-schulman15, otto2021differentiable}.
%
The crucial difference is that \ittr{} proposes a KL-bound in parameter space while the previous methods limit the KL-divergence in the output space.
Inspired by zero-order stochastic-search algorithms \citep{huettenrauch2022regret, Abdolmaleki2015, peters2010relative}, \citet{arenz2023a} propose a first-order method to directly optimize model parameters, but their approach does not gracefully scale to even small neural networks with thousands of parameters and even less so to larger ones. 


\section{Conclusion}
We introduced Information-Theoretic Trust Region Optimization (\ittr{}), a novel approach to stochastic optimization of deep neural networks. 
First, \ittr{} efficiently estimates second-order information from gradients using a recursive least squares approach.
In the parameter update, \ittr{} accounts for uncertainty in the second-order information and current parameters by limiting the updates using information-theoretic trust regions. 
\ittr{} matches the convergence behavior of approaches using gradient statistics for preconditioning and step size control, e.g., Adam \citep{Kingma2015}, while giving results comparable to those of a carefully tuned SGD with momentum.
In future work, we aim to use our approach for Bayesian Deep Learning. 
There are similarities between our objective and the variational evidence lower bound which we could further exploit and learn a full distribution over network parameters. 

\section{Acknowledgments}
This work was supported by funding from the pilot program Core Informatics of the Helmholtz Association (HGF). The authors acknowledge support by the state of Baden-Württemberg through bwHPC, as well as the HoreKa supercomputer funded by the Ministry of Science, Research and the Arts Baden-Württemberg and by the German Federal Ministry of Education and Research.

\newpage
\bibliography{bibliography}

\appendix
\section{Detailed solution of the Optimization problem} \label{sec:solution_opt_problem}
This section presents the detailed solution to the \ittr{} optimization problem
\begin{mini}|s| 
    {\pi_t}
    {\mathbb{E}_{\bm\theta \sim \pi_t}[f_t(\bm\theta)] + \rho C_\lambda + \nu C_\Sigma}{}{}
    \addConstraint{C_\mu}{\le \varepsilon}
\end{mini}
with
\begin{align*}
    \pi_t(\bm \theta) &= \mathcal{N}(\bm \theta; \bm \mu_t, \bm \Sigma_t) \\
    f_t(\bm \theta) &= \frac{1}{2} \bm \theta_t^T \bm A_t \bm \theta_t + \bm \theta^T\bm b_t + c_t, \\
    C_\lambda &= \kl(\pi_t || \mathcal{N}(0, \lambda^{-1} \bm I_n) \\
    C_\Sigma &= \frac{1}{2}\Big [ \tr(\bm \Sigma_{t-1}^{-1} \bm \Sigma_t) - n + \log \det(\bm \Sigma_{t-1}) - \log \det(\bm \Sigma_t)  \Big]\\
    C_\mu &= \frac{1}{2} (\bm \mu_t - \bm \mu_{t-1})^T \bm \Sigma_{t-1}^{-1}(\bm \mu_t - \bm \mu_{t-1}).
\end{align*}
The variables $\bm \mu_t, \bm \Sigma_t$ are the primal parameters. We define a dual parameter $\eta \in \RR$. Inserting the diagonal forms ${\bm \Sigma_t = \diag(\bm \sigma_t^2)}$ and $\bm A_t = \diag(\bm a_t)$, and solving the expectation over the quadratic surrogate,
the Lagrangian is given by
\begin{align*}
    L(\bm \mu_t, \bm \sigma_t^2, \eta) ={} &\frac 1 2 \sum_{j=1}^n\Big[\mu_{t,j}^2 a_{t,j} + a_{t,j} \sigma_{t,j}^2 + \mu_{t,j} b_{t,j} + c_t \Big] \\
    + \rho &\frac{1}{2}  \sum_{j=1}^n \Big[ \sigma_{t,j}^2\lambda -1 + \mu_{t,j}^2 \lambda + \log(\lambda^{-1}) - \log(\sigma_{t,j}^2) \Big] \\
    + \nu &\frac{1}{2} \sum_{j=1}^n \Big[\frac{\sigma_{t,j}^2}{\sigma_{t-1,j}^2} -1 + \log (\sigma_{t-1,j}^2) - \log(\sigma_{t,j}^2)\Big]\\
    + \eta &\frac{1}{2} \sum_{j=1}^n \frac{(\mu_{t,j} - \mu_{t-1,j})^2}{\sigma_{t-1,j}^2} - \eta \varepsilon.
\end{align*}
Here, we sum over the parameter dimensions $j=1,\dots,n$.
Next, we compute the derivative with respect to the mean and set it to zero
\begin{align*}
    \frac{\partial L}{\partial \mu_{t,j}} = \mu_{t,j} a_{t,j}  + b_{t,j} + \rho \lambda \mu_{t,j} + \eta \frac{\mu_{t,j} - \mu_{t-1,j}}{\sigma_{t-1,j}^2} \overset{!}{=} 0.
\end{align*}
This results in the primal solution of the mean

\begin{equation*}
        \bm \mu_t(\eta) = (\bm A_t + \eta \bm \Sigma_{t-1}^{-1} + \rho\lambda \bm I_n)^{-1}(\eta \bm \Sigma_{t-1}^{-1}\bm \mu_{t-1} - \bm b_t).
\end{equation*}
In the same way, we compute the primal solution of the variance
\begin{align*}
    \frac{\partial L}{\partial \sigma_{t,j}^2} &= \frac{1}{2} \Big[a_{t,j} + \rho \lambda + \frac{\rho}{\sigma_{t,j}^2}  + \frac{\nu}{\sigma_{t-1,j}^2} - \frac{\nu}{\sigma_{t,j}^2} \Big] \overset{!}{=} 0 \\
    \implies \quad \bm \Sigma_t &= (\rho + \nu)\Big(\bm A_t +\rho \lambda \bm I_n + \nu \bm \Sigma_{t-1}^{-1}\Big)^{-1}.
\end{align*}
Note, that $\bm \Sigma_t$ is independent of the Lagrange multiplier $\eta$ since we have dropped the covariance part of the KL divergence from the constraints.

It remains to solve the dual optimization problem to obtain $\eta$. The dual function is given by inserting the primal solution into the Lagrangian, resulting in
\begin{equation*}
    g(\eta) = L(\bm \mu_t(\eta), \bm \Sigma_t(\eta), \eta).
\end{equation*}
We solve the convex dual optimization problem
\begin{equation}
    \eta^* = \textrm{argmax}_{\eta} \;\; g(\eta), \quad \textrm{s.t. } \eta \ge 0 \label{obj:dual}
\end{equation}
by finding an $\eta^*$ with the derivative
\begin{equation*}
    g'(\eta^*) = 0.
\end{equation*}
The derivative of the dual has a simple form given by
\begin{equation}
    g'(\eta) = \frac{1}{2} (\bm \mu_t(\eta) -\bm \mu_{t-1})^T\bm \Sigma_{t-1}^{-1}(\bm \mu_t(\eta) - \bm \mu_{t-1}) - \varepsilon.
    \label{eq:dual_der}
\end{equation}
We find $\eta^*$ using a bisection method. It is possible that the derivative in  \cref{eq:dual_der} is always negative. This happens when the optimal solution lies inside the trust region. In that case, the solution to \cref{obj:dual} is given by $\eta^*=0$.

\section{Derivation of the Surrogate Model Fitting} 
\label{sec:drift_model}
This section describes the computation of the quadratic surrogate 
\begin{equation}
     \mathcal{L}(\bm \theta) \approx f_t(\bm \theta) = 0.5\bm \theta^T \bm A_t \bm \theta + \bm \theta^T \bm b_t + c_t \label{eq:quad_model}
\end{equation}

from previous gradient evaluations $\bm g_0=\nabla \mathcal{L}_0(\bm \theta)|_{\bm \theta = \bm \mu_0}, \dots, \bm g_t=\nabla \mathcal{L}_t(\bm \theta)|_{\bm \theta = \bm \mu_t}$. The general idea is to accumulate the information about the loss function into the current update using recursive least squares with a drift model. Taking the derivative on both sides of \cref{eq:quad_model}, the quadratic surrogate of the objective function is equivalent to a linear surrogate of its gradient
\begin{equation}
    \bm A_t \bm \theta + \bm b_t \approx \nabla_\theta \mathcal{L}(\bm \theta). \label{eq:linear_gradient_model}
\end{equation}

As we have seen in \cref{sec:main}, the update of the parameter distribution $\pi_t(\bm \theta)$ does not depend on the scalar parameter $c_t$ of the surrogate. Therefore, the linear surrogate of the gradient contains all the necessary information.
Selecting a diagonal matrix $\bm A_t = \diag(\bm a_t)$ leads to independent one-dimensional regression problems for fitting the surrogate.
We can find the values $a_t \in \RR$ and $b_t \in \RR$ from the gradients  $g_0, \dots, g_t \in \RR$ evaluated at the parameters $\mu_0, \dots \mu_t \in \RR$ for each dimension of the parameter space independently. To utilize matrix-vector formulations, we define the weight vector
\begin{equation*}
    \bm w_t =(a_t, b_t)^T.
\end{equation*}
A recursive least squares approach \citep{Saerkkae_2013} is a capable framework for solving this problem. In the recursive setting, we get one new observation and update our current belief accordingly.
The new observation is the gradient $g_t$ that we want to approximate. 
\begin{figure}
    \centering
        \begin{tikzpicture}[
    greennode/.style={circle, draw=black!80, fill=blue!7, very thick, minimum size=7mm},
    rednode/.style={circle, draw=black!80, fill=red!7, very thick, minimum size=7mm},
    invisible/.style={circle},
    ]
        \node[greennode] (x0) {$\bm{w_0}$} ;
        \node[rednode] (y0) [below=of x0]{$\bm{g_0}$};
        \node[greennode] (x1) [right=of x0] {$\bm{w_1}$};
        \node[rednode] (y1) [below=of x1]{$\bm{g_1}$};
        \node[greennode] (x2) [right=of x1] {$\bm{w_2}$};
        \node[rednode] (y2) [below=of x2]{$\bm{g_2}$};
        \node[greennode] (x3) [right=of x2] {$\bm{w_3}$};
        \node[rednode] (y3) [below=of x3]{$\bm{g_3}$};
        \node[invisible] (end) [right=of x3] {$\bm{\ldots}$};

        \draw[very thick,draw=black!80, ->] (x0) -- (x1);
        \draw[very thick,draw=black!80, ->] (x1) -- (x2);
        \draw[very thick,draw=black!80, ->] (x2) -- (x3);
        \draw[very thick,draw=black!80, ->] (x3) -- (end);
        
        \draw[very thick,draw=black!80, ->] (x0) -- (y0);
        \draw[very thick,draw=black!80, ->] (x1) -- (y1);
        \draw[very thick,draw=black!80, ->] (x2) -- (y2);
        \draw[very thick,draw=black!80, ->] (x3) -- (y3);
        
    \end{tikzpicture}
    \caption{Probabilistic state space model for fitting the quadratic surrogate. The drift of the surrogate parameters $\bm w_t$ over time is modeled using a Gaussian random walk. The observed gradients $g_t$ in each step are modeled using a linear model with Gaussian noise.}
    \label{fig:graphical_model}
\end{figure}
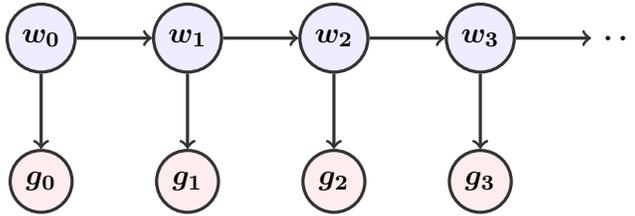

In general, there are two stochastic processes we have to model here. First, we allow the surrogate to drift over time since we are not evaluating the gradients at the same point in the parameter space throughout the optimization process. Therefore, older observations may not align with the current surrogate. Second, we have to deal with the fact that the gradient information is noisy due to the noisy objective  $\mathcal{L}_t(\bm \theta)$. This is a crucial detail since other second-order algorithms like L-BFGS~\citep{lbfgs} have suffered from this problem~\citep{lbfgs_fail_sgd}.
A formulation that addresses these issues builds upon a Bayesian view and utilizes a probabilistic state space model given in \cref{fig:graphical_model}. 
The goal is to compute the parameters $\bm m_t \in \RR^2$ and $\bm P_t \in \RR^{2 \times 2}$ of the filtering distribution
\begin{equation}
    p(\bm w_t| g_{0:t}) := \mathcal{N}(\bm w_t| \bm m_t, \bm P_t). \label{eq:filtering_distribution}
\end{equation}
We have no information about the development of $\bm w_t$ throughout the optimization. Hence, we model its dynamics as a Gaussian random  walk
\begin{equation*}
    p(\bm w_t|\bm w_{t-1}) = \mathcal{N}(\bm w_t | \bm w_{t-1}, q \bm I)
\end{equation*}
with drift variance $q \in \RR$.  To address the noisy gradient evaluations, 
we define a measurement model which describes the observation $g_t$ given the current state $\bm w_t$
\begin{equation*}
    p(g_t|\bm w_t) = \mathcal{N}(g_t | \bm H_t \bm w_t, r).
\end{equation*}
The matrix $\bm H_t = (\mu_t, 1) \in \RR^2$ displays the linear relationship of the gradient from \cref{eq:linear_gradient_model} while the value $r \in \RR$ models the noise of the gradients. The noise values $q$ and $r$ are hyperparameters for the \ittr{} algorithm.

Given a previous state space distribution  $p(\bm w_{t-1}|  g_{0 : t-1}) = \mathcal{N}(\bm w_{t-1}|\bm m_{t-1}, \bm P_{t-1})$, the update equation to obtain the filtering distribution \cref{eq:filtering_distribution} at step $t$ are the following \citep{Saerkkae_2013}
\begin{equation}
    \begin{split}
        \bm P_t^{-} &= \bm P_{t-1} +  q \bm{I}\\
        v_t &= \bm H_t \bm P_t^{-} \bm H_t^T + r\\
        \bm K_t &= \bm P_t^{-} \bm H_t^T v_t^{-1}\\
        \bm m_t &= \bm m_{t-1} + \bm K_t(g_t - \bm H_t \bm m_{t-1})\\
        \bm P_t &= \bm P_t^{-} - v_t \bm K_t \bm K_t^T.    
    \end{split} \label{eq:filter_update_rule}
\end{equation}

We use the maximum a posteriori solution 
$(a_t,b_t)^T := \bm m_t$
to approximate the parameters of the quadratic surrogate \cref{obj:itropt}. The described one-dimensional computation of $(a_t, b_t)^T$ from the current gradient evaluation $g_t$ shown in \cref{eq:filter_update_rule} is easily vectorizable and can be implemented efficiently even for high dimensional parameter spaces with millions of parameters. 


\section{Experiment Details} \label{sec:exp_details}
\label{app:exp_details}
\begin{figure*}[t]
\begin{minipage}{0.5\textwidth}
    \centering
    \includegraphics[width=\textwidth]{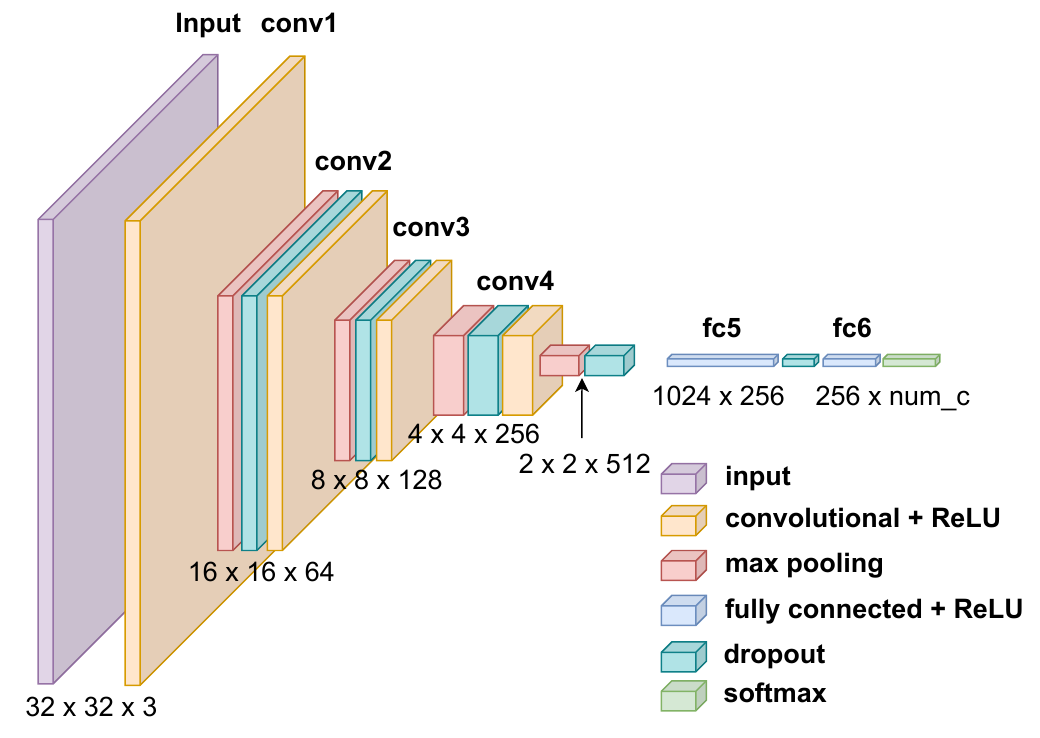}
\end{minipage}
\begin{minipage}{0.5\textwidth}
    \centering
    \includegraphics[width=\textwidth]{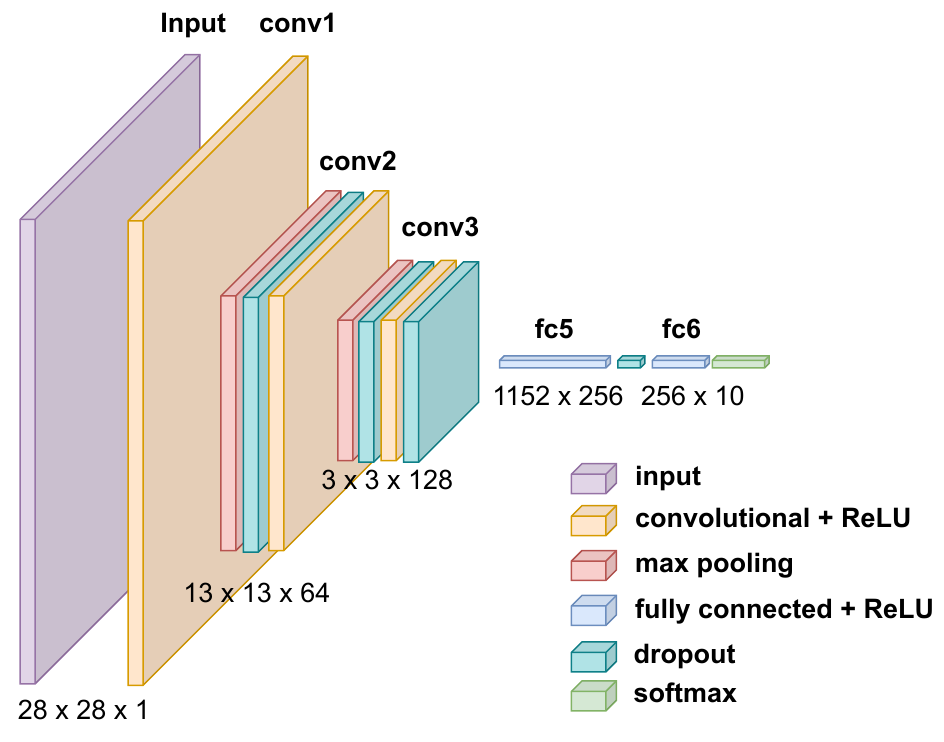}
\end{minipage}%
\caption{CNN architectures for the CIFAR-$10$/$100$ dataset (left) and  for the Fashion-MNIST dataset (right).}
\label{fig:cnn_arch}
\end{figure*}

This section lists the details for the empirical evaluation in Section 3. We present an illustration of the CNN architectures used for the experiments in \cref{fig:cnn_arch}. Regarding the ResNet architectures, we use the untrained architectures implemented by PyTorch in the torchvision package. For the Fashion-MNIST task, we change the first layer of the ResNet architecture to accept images with only one channel.

To obtain a fair comparison, we tune all hyperparameters for all algorithms on all available tasks. We use the Weights and Biases sweep functionality (Biewald, 2020). We apply k-fold cross-validation (Kohavi et al., 1995) with $k = 8$ for the HPO since the datasets
do not contain a separate validation set. For each configuration of task and algorithm, we test $50$ hyperparameter configurations and evaluate each configuration on $3$ different seeds. The seed has an impact on the initial weights of the architecture and the shuffling of the dataset. We use a Bayesian optimization (BO) which uses an initial distribution of hyperparameters and proposes new configurations of hyperparameters based on the validation accuracy of previous runs. After obtaining all the hyperparameter configurations per experiment, we cross-evaluated the combinations to see if another combination can perform better than the one found by the BO.

For \ittr{}, we have to define the trust region bound $\varepsilon$, the initial variance $\bm \Sigma_0$, the scaling values $\rho, \nu$ of the objective, the prior precision $\lambda$ and the noise values $r$, $q$ for fitting the surrogate. We use default values for the initial variance $\bm \Sigma_0 = 0.01$ of the parameter distribution, the prior precision $\lambda = 0.0015$, and the scaling $\nu=1.3$ of the covariance part of the KL since they did not have a big impact upon the results of the optimization. We further select a small value for the initial variance of the filtering distribution $\bm P_0 = \diag((0.00005, 0.00005))$.  The bisection method uses a warm start method for the interval bounds based on the previous computation of $\eta^*$. The lower and upper bound $C_l$ and $C_u$ are initialized as
\begin{align*}
    C_l &= \frac{\eta^*}{3},\\
    C_u &= 3 \eta^*.
\end{align*}
The bisection method either stops when 
\begin{equation*}
    C_u - C_l < 0.5,
\end{equation*}
or when
\begin{equation*}
    |g'(\eta)| < 0.1 \varepsilon.
\end{equation*}
This leaves us with four relevant hyperparameters, the trust region bound $\varepsilon$, the prior scaling $\rho$, and the noise values $r$ and $q$ required for the update of the surrogate model.

The tuned hyperparameters for all algorithms are given in \cref{tab:hp_itropt,tab:hp_sgd,tab:hp_adam,tab:hp_adamw}.

\begin{table}[]
    \centering
            \begin{tabular}{lcccccc}\toprule
        \multirow{ 2}{*}{\textbf{\large{\ittr{}}}}& \multicolumn{2}{c}{\textbf{Fashion-MNIST}} & \multicolumn{2}{c}{\textbf{CIFAR-10}} &
        \multicolumn{2}{c}{\textbf{CIFAR-100}} \\
                  & CNN  & ResNet-$18$ & CNN & ResNet-$34$  & CNN & ResNet-$34$ \\ 
      \midrule 
$\varepsilon$ & $0.085675$ & $0.085675$ & $0.002787$ & $0.007133$ & $0.002787$ & $0.007234$ \\ 
$\rho$ & $0.058657$ & $0.058657$ & $0.296786$ & $1.597354$ & $0.296786$ & $1.676770$ \\ 
$r$ & $2.816791$ & $2.816791$ & $1.219750$ & $9.328440$ & $1.219750$ & $4.822032$ \\ 
$q$ & $0.017393$ & $0.017393$ & $0.002455$ & $0.089381$ & $0.002455$ & $0.009779$ \\ 
Weight Decay & $0.000002$ & $0.000002$ & $0.000000$ & $0.000703$ & $0.000000$ & $0.000000$ \\ 
      \bottomrule
    \end{tabular}
    \caption{Tuned hyperparameter for the \ittr{} algorithm}
    \label{tab:hp_itropt}
\end{table}

\begin{table}[]
    \centering
            \begin{tabular}{lcccccc}\toprule
        \multirow{ 2}{*}{\textbf{\large{SGD}}}& \multicolumn{2}{c}{\textbf{Fashion-MNIST}} & \multicolumn{2}{c}{\textbf{CIFAR-10}} &
        \multicolumn{2}{c}{\textbf{CIFAR-100}} \\
                  & CNN  & ResNet-$18$ & CNN & ResNet-$34$  & CNN & ResNet-$34$ \\ 
      \midrule 
Learning Rate & $0.071049$ & $0.137031$ & $0.017834$ & $0.056480$ & $0.017834$ & $0.067994$ \\ 
Momentum & $0.865730$ & $0.854087$ & $0.946762$ & $0.866487$ & $0.946762$ & $0.867370$ \\ 
Weight Decay & $0.000225$ & $0.001963$ & $0.000163$ & $0.001697$ & $0.000163$ & $0.001800$ \\ 
      \bottomrule
    \end{tabular}
    \caption{Tuned hyperparameter for the SGD algorithm}
    \label{tab:hp_sgd}
\end{table}

\begin{table}[]
    \centering
            \begin{tabular}{lcccccc}\toprule
        \multirow{ 2}{*}{\textbf{\large{Adam}}}& \multicolumn{2}{c}{\textbf{Fashion-MNIST}} & \multicolumn{2}{c}{\textbf{CIFAR-10}} &
        \multicolumn{2}{c}{\textbf{CIFAR-100}} \\
                  & CNN  & ResNet-$18$ & CNN & ResNet-$34$  & CNN & ResNet-$34$ \\ 
      \midrule 
        Learning Rate & $0.001012$ & $0.045115$ & $0.001129$ & $0.006652$ & $0.001129$ & $0.001612$ \\ 
        $\beta_1$ & $0.945256$ & $0.907895$ & $0.851157$ & $0.890313$ & $0.851157$ & $0.864582$ \\ 
        $\beta_2$ & $0.990342$ & $0.999999$ & $0.998940$ & $0.999387$ & $0.998940$ & $0.999953$ \\ 
        Weight Decay & $0.000000$ & $0.000002$ & $0.001090$ & $0.000447$ & $0.001090$ & $0.001941$ \\ 
      \bottomrule
    \end{tabular}
    \caption{Tuned hyperparameter for the Adam algorithm}
    \label{tab:hp_adam}
\end{table}

\begin{table}[]
    \centering
            \begin{tabular}{lcccccc}\toprule
        \multirow{ 2}{*}{\textbf{\large{AdamW}}}& \multicolumn{2}{c}{\textbf{Fashion-MNIST}} & \multicolumn{2}{c}{\textbf{CIFAR-10}} &
        \multicolumn{2}{c}{\textbf{CIFAR-100}} \\
                  & CNN  & ResNet-$18$ & CNN & ResNet-$34$  & CNN & ResNet-$34$ \\ 
      \midrule 
        Learning Rate & $0.001004$ & $0.018744$ & $0.001129$ & $0.006652$ & $0.001129$ & $0.001245$ \\ 
        $\beta_1$ & $0.922247$ & $0.862748$ & $0.851157$ & $0.890313$ & $0.851157$ & $0.858643$ \\ 
        $\beta_2$ & $0.999945$ & $0.999999$ & $0.998940$ & $0.999387$ & $0.998940$ & $0.998802$ \\ 
        Weight Decay & $0.000142$ & $0.000040$ & $0.001090$ & $0.000447$ & $0.001090$ & $0.001804$ \\ 
      \bottomrule
    \end{tabular}
    \caption{Tuned hyperparameter for the AdamW algorithm}
    \label{tab:hp_adamw}
\end{table}

\section{Additional Results}
\label{app:additional_results}
\begin{figure*}[]
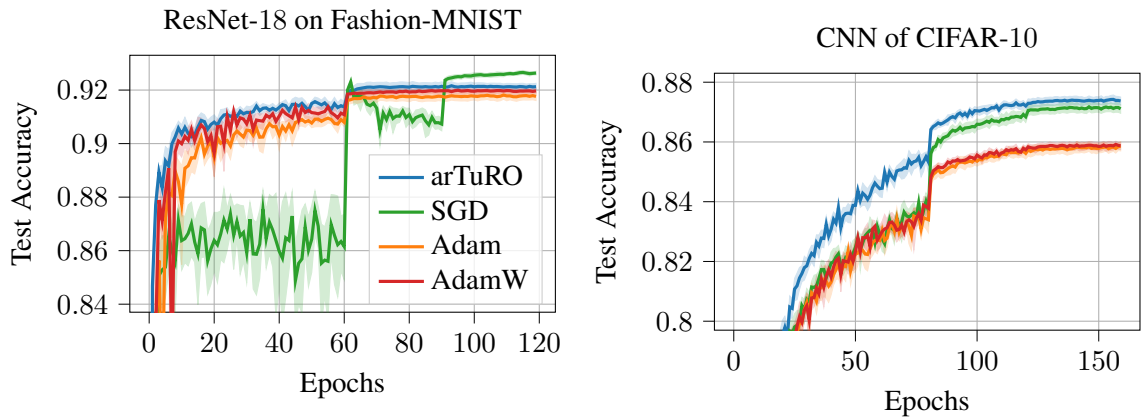

\begin{minipage}{0.5\textwidth}
\input{figures/results/fmnist_resnet18_test_acc}
\end{minipage}
\begin{minipage}{0.5\textwidth}
\centering
\input{figures/results/cifar10_dropout_cnn_test_acc}
\end{minipage}%
\caption{Test accuracies of the competing algorithms \ittr{}, SGD, Adam and AdamW on the two remaining experiments ResNet-$18$ on Fashion-MNIST (left) and CNN on CIFAR-$10$ (right). \ittr{} outperforms Adam and AdamW which is consistent to the other experiments. SGD beats \ittr{} on the ResNet-$18$ task applied to FashionMNIST, however its learning curve shows unstable behavior. }
\label{fig:add_test_acc}
\end{figure*}

\begin{figure*}[t]
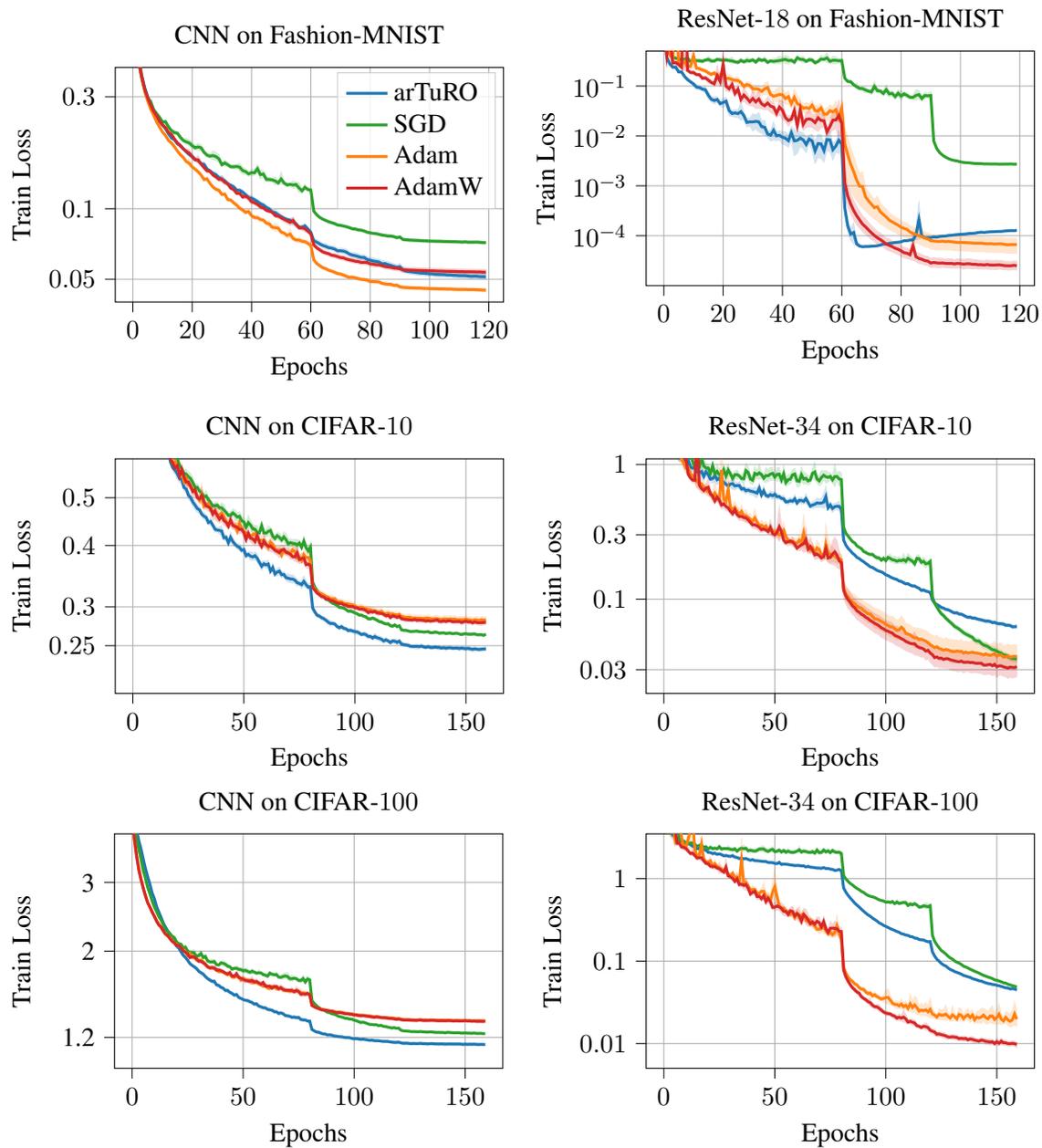

\begin{minipage}{0.5\textwidth}
\input{figures/train_losses/fmnist_dropout_cnn_train_loss}
\end{minipage}%
\begin{minipage}{0.5\textwidth}
\centering
\input{figures/train_losses/fmnist_resnet18_train_loss}
\end{minipage}
\begin{minipage}{0.5\textwidth}
\input{figures/train_losses/cifar10_dropout_cnn_train_loss}
\end{minipage}%
\begin{minipage}{0.5\textwidth}
\centering
\input{figures/train_losses/cifar10_resnet34_train_loss}
\end{minipage}
\begin{minipage}{0.5\textwidth}
\input{figures/train_losses/cifar100_dropout_cnn_train_loss}
\end{minipage}%
\begin{minipage}{0.5\textwidth}
\centering
\input{figures/train_losses/cifar100_resnet34_train_loss}
\end{minipage}
\caption{Train loss curves of the competing algorithms \ittr{}, SGD, Adam and AdamW.}
\label{fig:train_loss}
\end{figure*}

This section lists the missing figures of the test accuracies for the experiments ResNet-$18$ on Fashion-MNIST and CNN on CIFAR-$10$ in \cref{fig:add_test_acc}. We also present the train loss curves of all experiments in \cref{fig:train_loss}.

As mentioned in the paper, the \ittr{} algorithm needs a longer computation time compared to state-of-the-art optimizer. We present the average time per training epoch in \cref{tab:timing}. All experiments are run on a single NVIDIA GeForce RTX 3080.

\begin{table}[]
    \centering
\begin{tabular}{lcccccc}\toprule
        & \multicolumn{2}{c}{\textbf{Fashion-MNIST}} & \multicolumn{2}{c}{\textbf{CIFAR-10}} &
        \multicolumn{2}{c}{\textbf{CIFAR-100}} \\
                  & CNN  & ResNet-$18$ & CNN & ResNet-$34$  & CNN & ResNet-$34$ \\ 
      \midrule 
        SGD    & $1.23$ & $4.19$ & $6.60$ & $9.05$ & $7.09$ & $12.08$\\
        Adam   & $1.33$ & $5.16$ & $6.71$ & $9.83$ & $7.21$ & $11.33$\\
        AdamW  & $1.28$ & $5.13$ & $6.80$ & $9.92$ & $7.18$ & $11.57$\\
        \ittr{} & $1.85$ & $5.42$ & $7.40$ & $15.59$ & $7.92$ & $15.87$\\
      \bottomrule
    \end{tabular}

    \caption{Average time per training epoch in seconds. }
    \label{tab:timing}
\end{table}

\end{document}